\documentclass[letterpaper, 10 pt, conference]{ieeeconf}
\IEEEoverridecommandlockouts
\usepackage[dvipsnames]{xcolor}

\title{\LARGE \bf
Swashplateless-elevon Actuation for a Dual-rotor Tail-sitter VTOL UAV
}

\author{Nan Chen$^{*}$, Fanze Kong$^{*}$, Haotian Li, Jiayuan Liu, Ziwei Ye, \\Wei Xu, Fangcheng Zhu, Ximin Lyu, and Fu Zhang
\thanks{$^*$These two authors contribute equally to this work.}
\thanks
{N. Chen, F. Kong, H. Li, J. Liu, Z. Ye, F. Zhu, W. Xu, and F. Zhang are with Department of Mechanical Engineering, The University of Hong Kong. \{\textit{cnchen,kongfz,haotianl,liujy000,yiziwai,zhufc}\}\textit{@connect.hku.hk}, \{\textit{xuweii,fuzhang}\} \textit{@hku.hk}. X. Lyu is with School of intelligent systems engineering, Sun Yat-sen University, Shenzhen. \textit{lvxm6@mail.sysu.edu.cn}}
}

\usepackage{epsfig} % for postscript graphics files
\usepackage{times} % assumes new font selection scheme installed
\usepackage{amsmath} % assumes amsmath package installed
\usepackage{amssymb} % assumes amsmath package installed
\usepackage{amsmath,amssymb,amsopn,amstext,amsfonts}
\usepackage{bm}
\usepackage{float}
\usepackage{booktabs}
\usepackage{cite} %for realize [1]-[3]

\usepackage{moreverb,url}

\begin{document}

\maketitle
% \thispagestyle{plain}
% \pagestyle{plain}

%%%%%%%%%%%%%%%%%%%%%%%%%%%%%%%%%%%%%%%%%%%%%%%%%%%%%%%%%%%%%%%%%%%%%%%%%%%%%%%%
\begin{abstract}

In this paper, we propose a novel swashplateless-elevon actuation (SEA) for dual-rotor tail-sitter vertical takeoff and landing (VTOL) unmanned aerial vehicles (UAVs). In contrast to the conventional elevon actuation (CEA) which controls both pitch and yaw using elevons, the SEA adopts swashplateless mechanisms to generate an extra moment through motor speed modulation to control pitch and uses elevons solely for controlling yaw, without requiring additional actuators. This decoupled control strategy mitigates the saturation of elevons' deflection needed for large pitch and yaw control actions, thus improving the UAV's control performance on trajectory tracking and disturbance rejection performance in the presence of large external disturbances. Furthermore, the SEA overcomes the actuation degradation issues experienced by the CEA when the UAV is in close proximity to the ground, leading to a smoother and more stable take-off process. We validate and compare the performances of the SEA and the CEA in various real-world flight conditions, including take-off, trajectory tracking, and hover flight and position steps under external disturbance. Experimental results demonstrate that the SEA has better performances than the CEA. Moreover, we verify the SEA's feasibility in the attitude transition process and fixed-wing-mode flight of the VTOL UAV. The results indicate that the SEA can accurately control pitch in the presence of high-speed incoming airflow and maintain a stable attitude during fixed-wing mode flight. Video of all experiments can be found in \url{youtube.com/watch?v=Sx9Rk4Zf7sQ}

\end{abstract}

%%%%%%%%%%%%%%%%%%%%%%%%%%%%%%%%%%%%%%%%%%%%%%%%%%%%%%%%%%%%%%%%%%%%%%%%%%%%%%%%
\section{INTRODUCTION}

Vertical takeoff and landing (VTOL) unmanned aerial vehicles (UAVs) can take off and land vertically like multi-rotor UAVs and achieve efficient long-range and high-speed flight similar to fixed-wing UAVs \cite{gandhi2019practical}. VTOL UAVs can be implemented using different configurations, such as tilt-rotor \cite{chen2019design, cardoso2021new}, tilt-wing \cite{rohr2019attitude, sanchez2020development}, and tail-sitter \cite{lyu2017design, xu2019full, li2018model}. Among these, the tail-sitter UAV is advantageous due to its ability to transition its attitude to enter fixed-wing flight mode without additional tilting mechanisms, resulting in a simpler and more compact structure \cite{xu2020learning}.

Tail-sitter UAVs can be controlled using various actuator configurations, such as quad-rotor type \cite{oosedo2017optimal, li2018development} and dual-rotor type with additional control surfaces (e.g., elevons) \cite{sun2018design, tal2021global, stone2008flight}. Compared with the quad-rotor type, the dual-rotor type uses fewer motors and propellers, making it less expensive, lighter, and more portable. However, the limited deflection range and small size of the additional control surfaces may lead to saturation, which constrain the achievable performance or even cause instability \cite{zhong2019l}. In particular, when the UAV hovers in multi-rotor mode, wind disturbances can significantly affect the UAV's attitude due to its large wing area. These disturbances often act on the pitch and yaw directions of the dual-rotor-type VTOL UAVs, which are controlled by the elevons (call it conventional elevon actuation (CEA)). One problem of the CEA is that large control efforts in both directions can induce elevons' deflection sharply, causing actuation saturation that further degrades the control performance or even destabilizes the system. One approach to mitigate this problem is to place elevons at the high-speed airflow region of the propellers by reducing the distance between rotors and elevons \cite{wang2019design}. However, this design also decreases the airflow speed flowing through the main wing, resulting in lift loss. Overall, wind-resistance is one of the major challenges in moving the dual-rotor tail-sitter VTOL UAV towards practical use.

\begin{figure}[t]
    \begin{center}
        {\includegraphics[width=1\columnwidth]{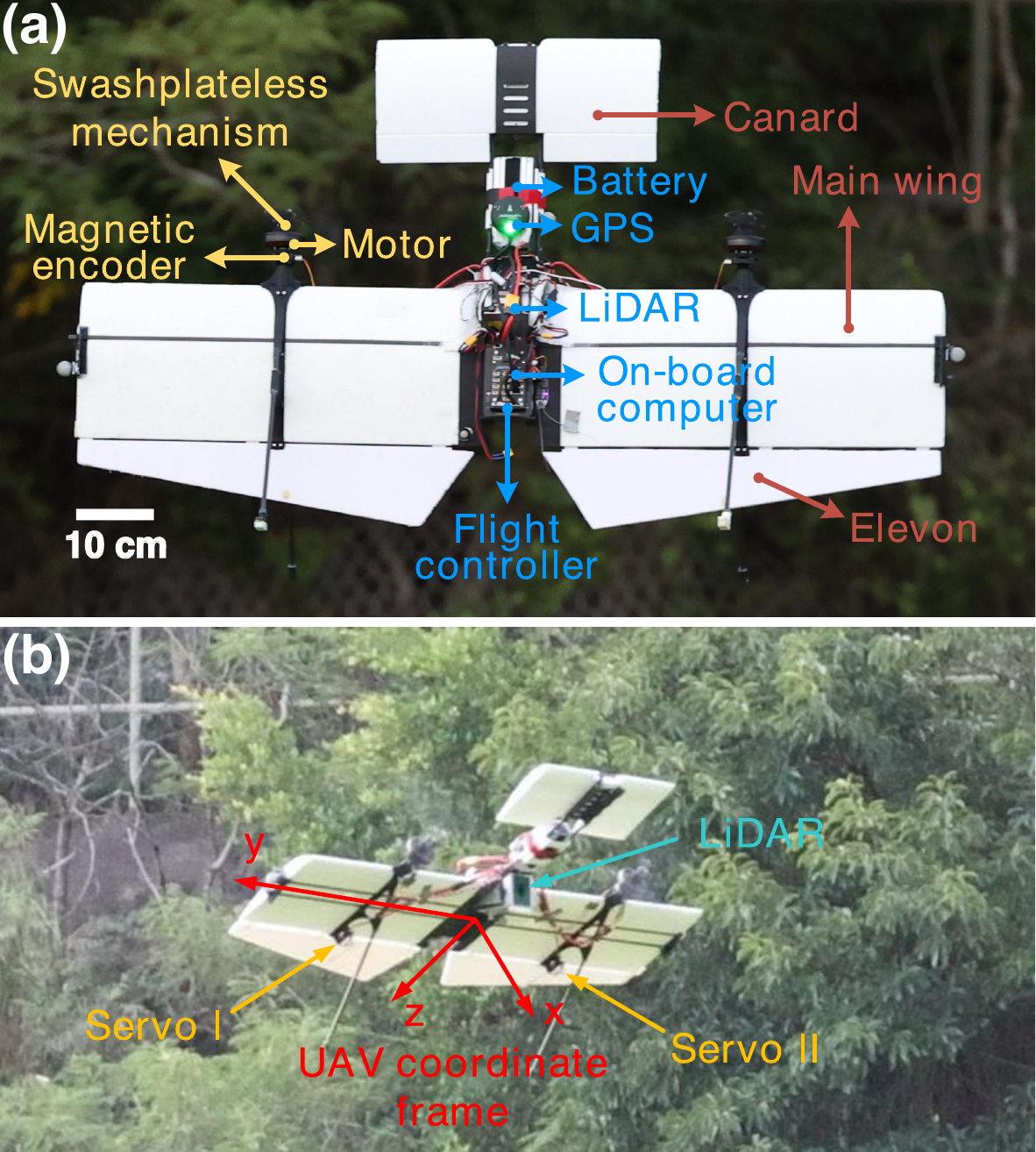}}
    \end{center}
    \vspace{-0.3cm}
    \caption{Dual-rotor tail-sitter VTOL UAV, named Hong He. (a) The UAV hovers at the multi-rotor mode. Main components on the front side of the UAV are labeled. (b) The UAV flies at the fixed-wing mode. Main components on the back side of the UAV and the definition of coordinate frame are shown.}
    \label{fig:cover}
    \vspace{-0.5cm}
\end{figure}

Another problem of CEA is encountered in the take-off process. Due to the constraints of the UAV footprint and landing stability, the landing gear cannot be very high, resulting in a short distance between the elevons and the ground. Before take-off, the downwash airflow generated from the propellers can be easily affected by the ground, leading to a decreased or even disappeared control moment of the elevons \cite{baker1970flight}. As a result, the pitch attitude may become unstable, potentially causing the UAV to crash.

Motivated by these problems, in this study, we propose a swashplateless-elevon actuation (SEA) which employs a swashplateless mechanism with improved structure to dual-rotor tail-sitter VTOL UAVs. The swashplateless mechanism is a passive structure mounted on the motors. It generates lateral moment by controlling the high-frequency acceleration and deceleration of the motor without requiring additional actuators \cite{paulos2015flight, paulos2018cyclic}. For VTOL UAVs, it can produce an extra control moment in the pitch direction, which can dramatically alleviate elevons' deflection saturation caused by large disturbances applied in both pitch and yaw directions or by large moments required by controllers, leading to a better disturbance rejection performance and higher maneuverability. Furthermore, the moment generated by the swashplateless mechanism is independent of the propeller's airflow, which enables a more stable take-off process for this type of UAV. We compare the SEA with the CEA in various experiments, which are described in the following sections.

\section{SYSTEM DESIGN}

\subsection{UAV Structure and Avionics}

We designed and manufactured a dual-rotor tail-sitter VTOL UAV named Hong He as shown in Fig. \ref{fig:cover}. For realizing large-scale scanning and mapping, it is equipped with a Livox AVIA LiDAR and an onboard computer Manifold 2-C. The LiDAR is capable of achieving long-distance scanning up to 400 m, making it suitable for outdoor large-scale scenes. To meet the requirements of long-range flight and control performance, we choose a canard wing layout and optimize the aircraft's aerodynamic shape design. For this prototype, our mechanical design concept emphasizes ease of assembly and maintenance, and hence we use formed wings and 3D printed materials (e.g., PLA, Nylon) to manufacture the UAV, making it simple to assemble.

The avionics control system is centered around a Pixhawk mini 4 flight controller attached to the fuselage. It is connected to two Electronic Speed Controllers (ESCs) (model T-MOTOR F35A 3-6S) for driving two motors (model T-MOTOR MN5006 KV450), two servos (model KST DS215MG V3.0) to control the deflection of two elevons, and two magnetic encoders (model AS5600) to measure the rotor angles of two motors. The final designed aircraft has a wingspan of 107 cm, weighs about 2.25 kg, and has a cruising speed of approximately 9 m/s. Detailed parameters can be found in Table \ref{hardware parameters}.

As shown in Fig. \ref{fig:cover} (b), the origin of UAV's coordinate frame (i.e. body frame) is attached to the UAV's center of mass. Beside the $x$ axis is perpendicular to the plane of main wing, both $y$ and $z$ axes are located in the plane of main wing, but the $y$ axis is parallel with the main wing while the $z$ axis points to the tail of UAV. The roll, pith, and yaw are defined in UAV's multi-rotor mode, and hence their rotation are along with the $x$, $y$, and $z$ axes, respectively. All the experimental data in this paper follow this definition.

\subsection{Swashplateless Mechanism}

\begin{figure}[t]
    \begin{center}
        {\includegraphics[width=1\columnwidth]{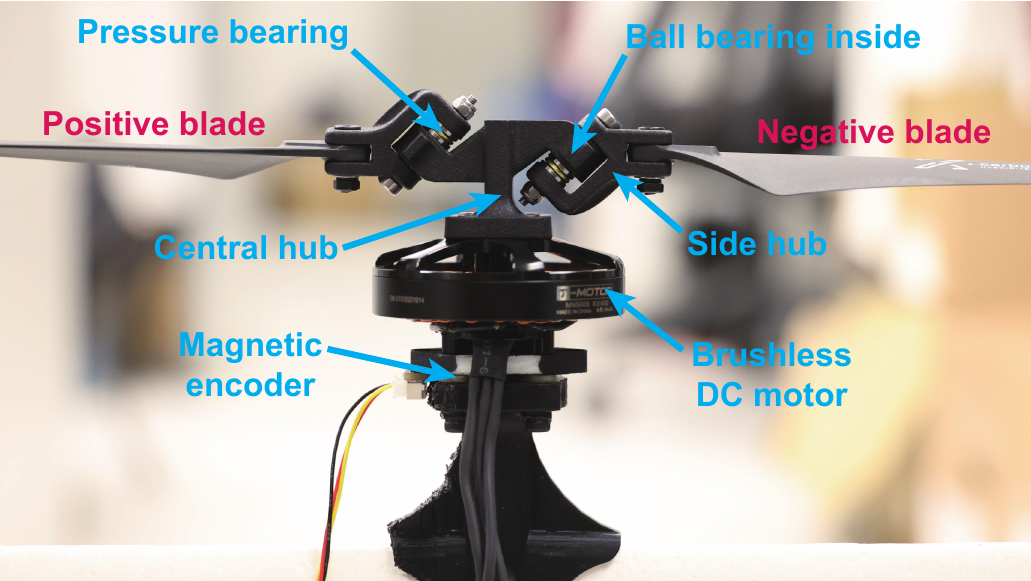}}
    \end{center}
    \vspace{-0.3cm}
    \caption{Components of one set of Hong He's propulsion system, mainly including a motor, a swashplateless mechanism, and two blades.}
    \label{fig:swash}
    \vspace{-0.5cm}
\end{figure}

The swashplateless mechanism, originally proposed in \cite{paulos2013underactuated}, can realize functions similar to the cyclic blade pitch control mechanism of swashplates that has been widely used in helicopters, providing both thrust and moment. The moment of the swashplateless mechanism comes from the unbalanced thrust of blades, induced by cyclic blade pitch changes. Unlike traditional swashplates driven by additional servos, the swashplateless mechanism is entirely passive. As shown in Fig. \ref{fig:swash}, two passive hinges connect side hubs to the central hub asymmetrically. Through periodic acceleration and deceleration of the motor (i.e., motor speed modulation), the unsymmetrical hinges rotate due to blade inertia, leading to different pitch angle changes of positive and negative blades. The blade with an increased pitch angle produces more thrust, while the blade with a decreased pitch angle produces less thrust, resulting in a net moment being generated. Detailed working principles are detailedly introduced in \cite{paulos2013underactuated,chen2023self,qin2022gemini}.

The propulsion system shown in Fig. \ref{fig:swash} mainly composes of a brushless DC motor, an improved swashplateless mechanism mounted on top of the motor's rotor, a pair of propeller blades, and a magnetic encoder mounted on the bottom of the motor's shaft. Compared with the original design, the improved structure of swashplateless mechanism includes additional ball bearings and pressure bearings to reduce the friction caused by the high-frequency rotation of the hinges, providing a smoother and more linear moment output and a faster response.

\begin{table}[t]
\caption{Specifications of the VTOL UAV Hong He.}
\label{hardware parameters}
\centering
\vspace{-0.2cm}
\begin{tabular}{cc}
\toprule[1.8pt]
Item           & Parameter \\
\toprule[1pt]
Wing span           & 107 cm \\
Total weight (with battery) & 2.25 kg \\
Thrust weight ratio & 2.5     \\
Cruise speed        & 9 m/s   \\
Cruise current      & 9 A    \\
Hover current       & 12 A    \\
Battery             & 6S, 2600 mAh  \\ \hline
\toprule[1.8pt]
\vspace{-0.5cm}
\end{tabular}
\end{table}

\begin{figure*}[t]
    \begin{center}
        {\includegraphics[width=2\columnwidth]{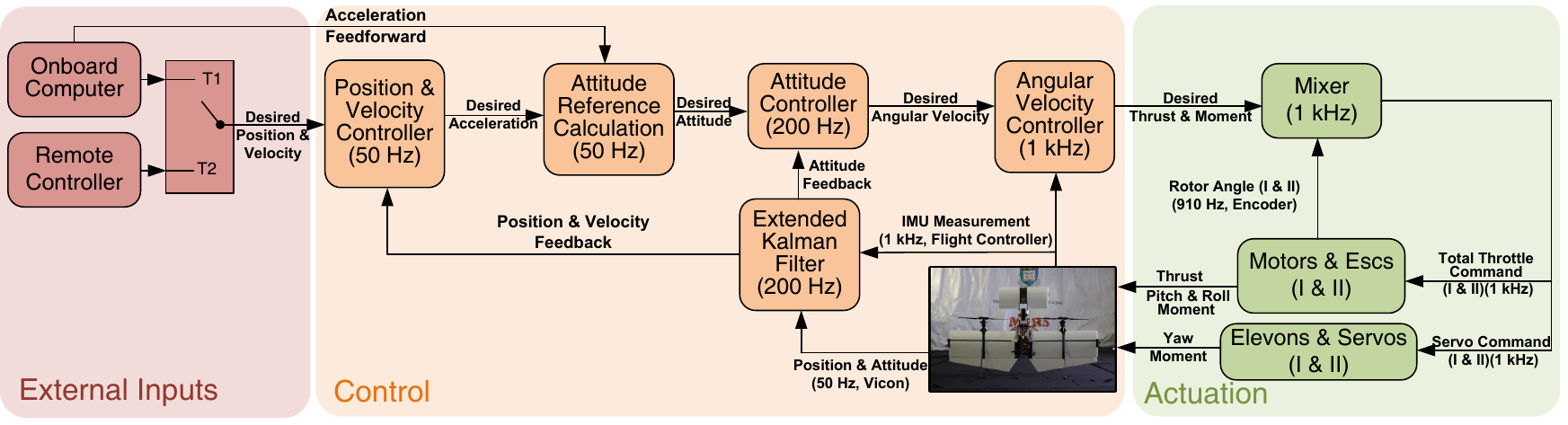}}
    \end{center}
    \vspace{-0.3cm}
    \caption{Control system overview. All modules in the control part are inherited from the standard PX4 and the mixer module of actuation part is additionally implemented into the PX4.}
    \label{fig:software}
    % \vspace{-0.3cm}
\end{figure*}

\section{Control and Actuation}
\subsection{Overview of Control System}

The overview of Hong He's control system with the proposed SEA is shown in Fig. \ref{fig:software}. The external inputs of the position \& velocity controller can be switched to two sources: remote controller (i.e., manual mode) and onboard computer (i.e., autonomous mode). In the manual mode, the remote controller directly generates the desired velocity. In the autonomous mode, the onboard computer produces desired position as the input of the position controller and the desired velocity \& acceleration as the feed-forward terms.

The control section (i.e., the orange zone) is the standard realization of PX4 while the actuation section (i.e., the green zone) are specifically designed for actuating the swashplateless mechanism and hence should be focused on. The operating frequency of the SEA mixer is dependent on the output frequency of the angular velocity controller (i.e., the measurement frequency of the inertial measurement unit (IMU)). Therefore, we set the measurement frequency of the IMU to 1 kHz, allowing the mixer run at the same frequency for processing the 910-Hz measurement from the two magnetic encoders without dropping any measurement. The outputs of the mixer are motor commands and servo commands, which are sent to the ESCs using the DShot600 communication protocol and to the servo by a standard 50-Hz PWM signal. The DShot600 protocol has a very short communication delay of 26.7 us, making it suitable for high-frequency speed modulation of the motor.

\subsection{Actuation Principles of SEA and CEA}

 The SEA and CEA exhibit identical actuation principles with respect to thrust generation and the control of roll and yaw moments. The two propellers, which are driven by motors, contribute to the total UAV thrust. Differential thrust produced by the propellers induces the roll control moment, while the airflow deflection by the elevons create the yaw control moment. However, there exists a difference between the SEA and CEA in terms of pitch control moment generation. The elevons contribute to the pitch control moment in the CEA, whereas the SEA employs swashplateless mechanisms for this purpose. This actuation principle facilitates decoupling of the attitude control of pitch and yaw.

In the CEA, the mixer blends the pitch and yaw moments to regulate the elevons' deflection through servo commands. Conversely, in the realization of SEA (i.e., the green zone in Fig. \ref{fig:software}), the pitch and yaw controls are directed towards separate actuators (i.e., the motors and the elevons). The SEA can mitigate elevon saturation by decoupling the generation of pitch control moment, thereby enhancing control and disturbance rejection performance. This feature is of particular significance, given that disturbances are frequently applied in the pitch and yaw directions due to the UAV's large main wing surface area.

\subsection{Design of SEA Mixer}

\subsubsection{Cyclic speed control of motor}

\begin{figure}[t]
    \begin{center}
        {\includegraphics[width=1\columnwidth]{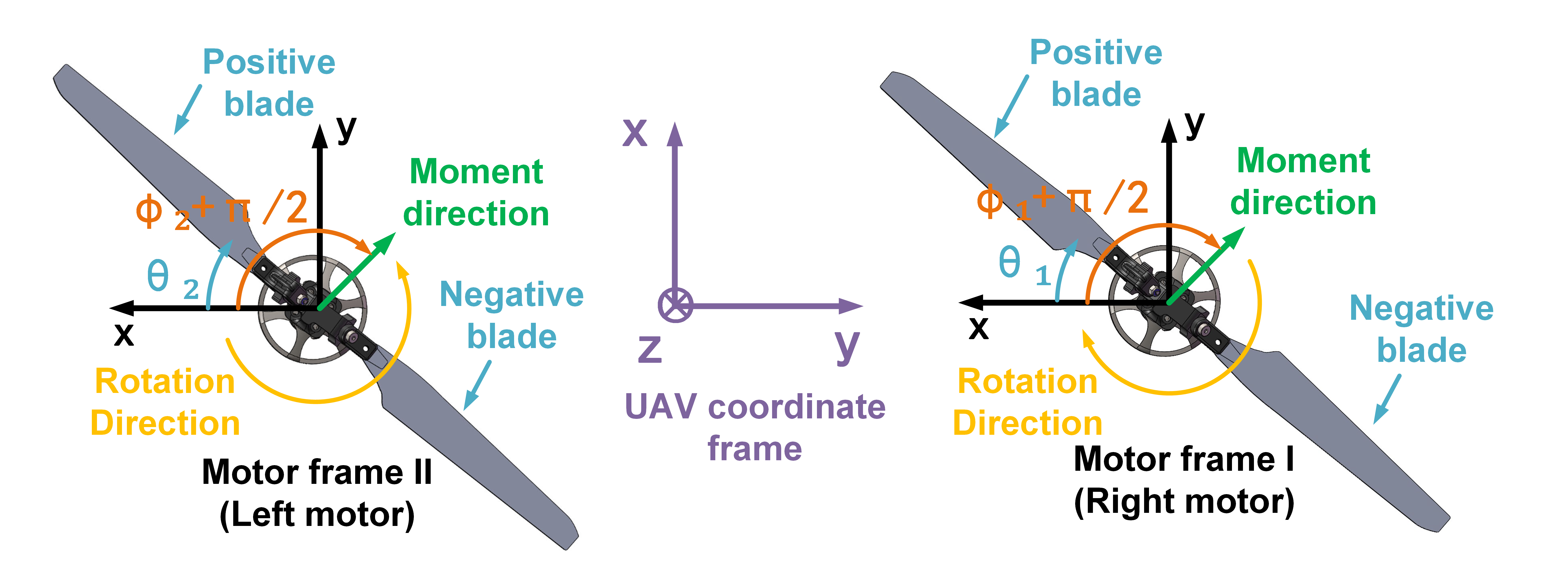}}
    \end{center}
    \vspace{-0.3cm}
    \caption{Coordinate frame definition for the swashplateless mechanism of the two motors shown in the top view of the UAV. The UAV coordinate frame is also labeled for reference.}
    \label{fig:motor_frame}
     % \vspace{-0.5cm}
\end{figure}

Before introducing the mixer, the cyclic speed control of motor is needed to be described since it defines a part of the output variables of the mixer. To prevent undesired vibration caused by motor speed modulation, a sinusoidal signal is employed for two motors. The total motor throttle $U_i, i=1,2$ is designed as
\begin{equation} \label{eq:sin_throttle}
\begin{aligned}
\left[ \begin{matrix}
    U_{1}          \\
    U_{2}
    \end{matrix} \right]
    =
    \left[ \begin{matrix}
    C_1 \\
    C_2
    \end{matrix} \right]
    +
    \left[ \begin{matrix}
    A_1 & 0 \\
    0 & A_2
    \end{matrix} \right]
    \left[ \begin{matrix}
    \vspace{0.1cm}
    \cos(\theta_1 - \phi_1 + \gamma_0) \\
    \cos(\theta_2 - \phi_2 - \gamma_0)
    \end{matrix} \right],
\end{aligned}
\end{equation}
where $C_i$, $A_i$, $\theta_i$, and $\phi_i$ are the nominal throttle, sinusoidal amplitude, motor's rotor angle measured by the magnetic encoder, and the moment direction in the UAV coordinate frame of motor $i$, $i=1,2$, respectively, the $\gamma_0$ is a positive constant for compensating the angle delay caused by blades' inertia and can be calibrated in advance through experiment data from a test stand of the swashplateless mechanism.

Despite the fact that the motors rotate in opposite directions, we can define the same coordinate frame for both motors, as shown in Fig. \ref{fig:motor_frame}. The absolute angle of the rotor, $\theta_i$, is defined as the angle between the $x$-axis and the positive blade, regardless of the rotation direction. In the motor coordinate frame, the moment direction is represented as $\phi_i + \pi/2$. However, due to the 90$^{\circ}$ rotation between the UAV frame and the motor frame, $\phi_i$ can represent the actual moment direction in the UAV frame. The sign of $\gamma_0$ is determined by the rotation direction since the delay direction is opposite to the rotation direction. As only the pitch angle is controlled by the moment of the swashplateless mechanism, $\phi_i$ can be simply set as $0$ or $\pi$, depending on the direction of the desired pitch moment.

\subsubsection{Mixer mapping}
The mixer is designed to calculate all actuator inputs (e.g., motor throttles, servo angles) according to the desired thrust $f_{T,d}^{\mathcal{B}}$ and moment $\bm \tau^\mathcal{B}_d$. Given the geometry of the
mechanical structure of the UAV, the mapping from actuator outputs to the body thrust and moment are
\begin{equation} \label{eq:mix_1}
    \begin{aligned}
    \left[ \begin{matrix}
    \vspace{0.1cm}
    f_{T}^{\mathcal{B}} \\
    \vspace{0.1cm}
    \tau_{x}^{\mathcal{B}} \\
    \vspace{0.1cm}
    \tau_{y}^{\mathcal{B}} \\
    \tau_{z}^{\mathcal{B}}
    \end{matrix} \right]
    =
    \left[ \begin{matrix}
    1 & 1  & 0 & 0 & 0 & 0     \\
    -L & L & 0 & 0 & 0 & 0     \\
    0 & 0  & 1 & 1 & 0 & 0 \\
    0 & 0  & 0 & 0 & 1 & 1 \\
    \end{matrix} \right]
    \left[ \begin{matrix}
    T_1          \\
    T_2          \\
    \tau_{s,1}          \\
    \tau_{s,2}          \\
    \tau_{e,1}     \\
    \tau_{e,2}
    \end{matrix} \right],
    \end{aligned}
\end{equation}
where $T_i$ and $\tau_{s,i}$ are the thrusts and moments generated by the motor $i, i=1,2$, respectively. $\tau_{e,j}$ are the moments generated by the elevon $j, j=1,2$, respectively. $L$ is the body-$y$-axis distance between UAV's center of mass and the motor.

Assuming that the actuator outputs are approximately linear to the actuator inputs, leads to
\begin{equation} \label{eq:mix_2}
    \begin{aligned}
    &\left[ \begin{matrix}
    T_1,
    T_2,
    \tau_{s,1},
    \tau_{s,2},
    \tau_{e,1},
    \tau_{e,2}
    \end{matrix} \right]^T
    =
    \\
    & \text{diag}(K_t, K_t, K_a, K_a, K_e, K_e)
    \left[ \begin{matrix}
    C_1,
    C_2,
    A_1,
    A_2,
    \delta_1,
    \delta_2
    \end{matrix} \right]^T,
    \end{aligned}
\end{equation}
where $\delta_j$ are the angle command of servo $j$, $j=1,2$. $K_t$, $K_a$, and $K_e$ are proportion coefficients, representing the conversions from throttle to thrust, from sinusoidal amplitude to swashplateless mechanism moment, and from servo angle to elevon moment, respectively. Here, the $C_1$, $C_2$, $A_1$, and $A_2$ are required by the cyclic speed control of motor.

Given the desired thrust $f_{T,d}^{\mathcal{B}}$ and the moment vector $\bm \tau_d^\mathcal{B}=\left[\begin{matrix}\tau_{x,d}^{\mathcal{B}} &\tau_{y,d}^{\mathcal{B}} & \tau_{z,d}^{\mathcal{B}}\end{matrix}\right]^T$, combining (\ref{eq:mix_1}) and (\ref{eq:mix_2}) and assuming the equal distribution of moments, we can finally determine the desired actuator inputs as
\begin{equation} \label{eq:mix_3}
\begin{aligned}
\left[ \begin{matrix}
    C_{1,d}          \\
    C_{2,d}          \\
    A_{1,d}          \\
    A_{2,d}          \\
    \delta_{1,d}     \\
    \delta_{2,d}
    \end{matrix} \right]
    =
    \left[ \begin{matrix}
    \vspace{0.1cm}
    \frac{1}{2 K_t} & \frac{-1}{2 L K_t} & 0 & 0 \\\vspace{0.1cm}
    \frac{1}{2 K_t} & \frac{1}{2 L K_t} & 0 & 0  \\\vspace{0.1cm}
    0 & 0  & \frac{1}{2K_a} & 0 \\\vspace{0.1cm}
    0 & 0  & \frac{1}{2K_a} & 0 \\\vspace{0.1cm}
    0 & 0  & 0 & \frac{1}{2K_e}  \\\vspace{0.1cm}
    0 & 0  & 0 & \frac{1}{2K_e}
    \end{matrix} \right]
    \left[ \begin{matrix}
    \vspace{0.1cm}
    f_{T,d}^{\mathcal{B}} \\
    \vspace{0.1cm}
    \tau_{x,d}^{\mathcal{B}} \\
    \vspace{0.1cm}
    \tau_{y,d}^{\mathcal{B}} \\
    \tau_{z,d}^{\mathcal{B}}
    \end{matrix} \right],
\end{aligned}
\end{equation}
where the $\delta_{j,d}$ are directly sent to the servo $j$ by PWM signal. The $C_{i,d}$ and $A_{i,d}$ are used to generate the total motor throttle $U_{i,d}$ sent to the ESCs. In (\ref{eq:mix_3}), the desired moments of pitch and yaw are decoupled into different actuators (i.e., the motors' swashplateless mechanisms and elevons).

\section{Experiments}

This section presents experimental results comparing the performance of the SEA and the CEA in several aspects, including take-off, trajectory tracking, and disturbance rejection. Additionally, tests of attitude transition and fixed-wing-mode flight are also performed using the SEA. To ensure a fair comparison, the controllers' parameters of angular rate, attitude, and position are tuned carefully in the best effort to optimize the performance for both SEA and CEA. The tuning process are conducted by analyzing the logged actual flight data to ensure fast responses while keeping small overshoots. All experiments can be found in an accompanying video uploaded in \url{youtube.com/watch?v=Sx9Rk4Zf7sQ}

\subsection{Comparison of Take-off Performance}

\begin{figure}[t]
    \begin{center}
        {\includegraphics[width=1\columnwidth]{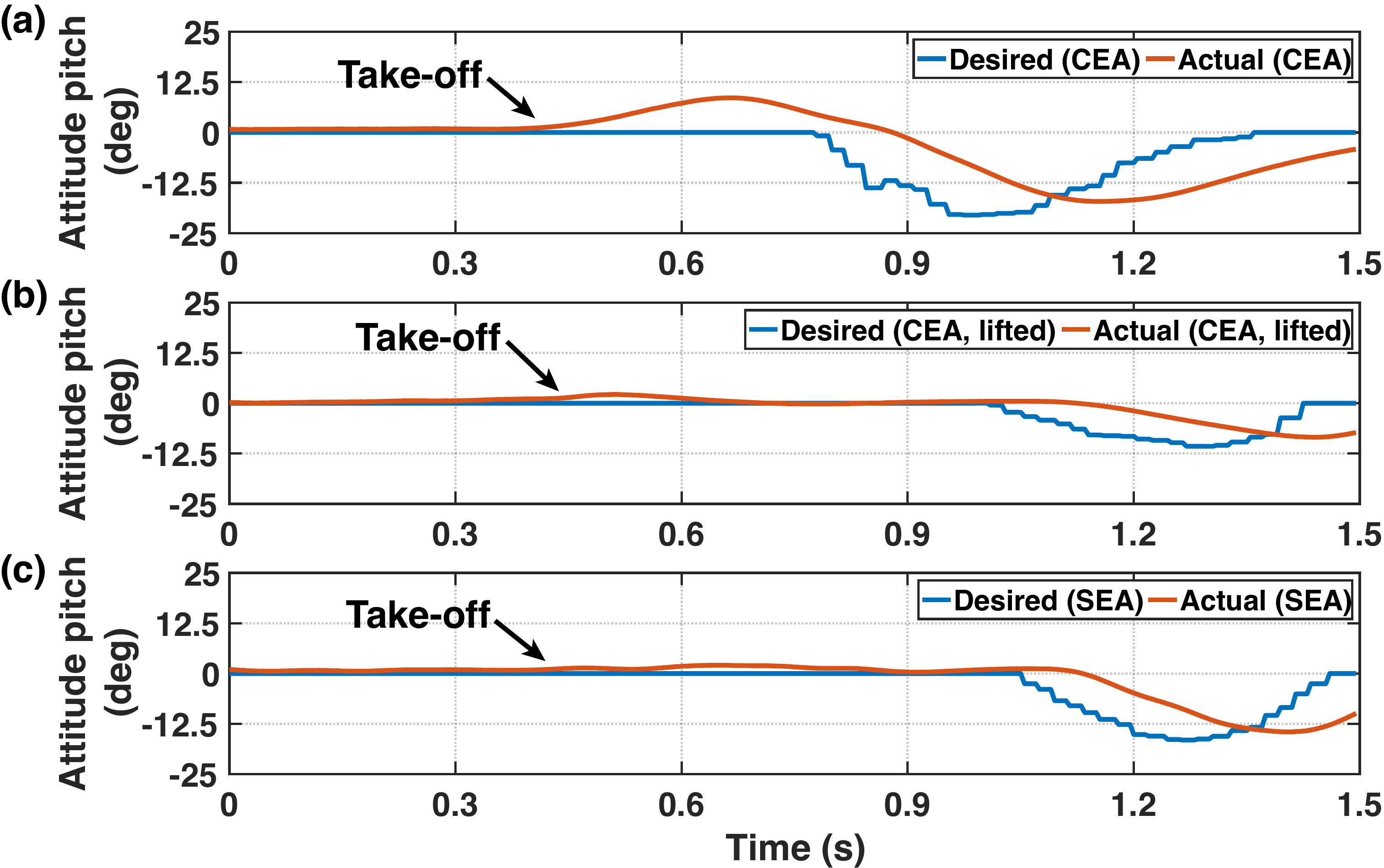}}
    \end{center}
    \vspace{-0.3cm}
    \caption{Comparison of pitch angle error during take-off in attitude control mode with different conditions.}
    \label{fig:take-off-exp}
\end{figure}

First, three take-off experiments are conducted on the UAV with attitude control (i.e., temporarily turn off the position controller). They are CEA with ground take-off, CEA with take-off on a lifted pedestal, and SEA with ground take-off. The results are shown in Fig. \ref{fig:take-off-exp} (a)-(c), respectively. In Fig. \ref{fig:take-off-exp} (a), the pitch angle unexpectedly increases to a maximum value of 8.6$^{\circ}$, although the desired attitude is still zero during take-off. The large pitch error during ground take-off is caused by the small distance between elevons and the ground. Constrained by both the footprint and landing stability, the landing gear cannot be higher, making the elevons easily affected by the near-ground airflow and deteriorating their control effect. In contrast, Fig. \ref{fig:take-off-exp} (b) shows the same take-off process on a lifted pedestal, and the pitch error significantly decreases to a maximum value of 2.1$^{\circ}$. The lifted pedestal alleviates the effect caused by the near-ground airflow, enabling the elevons to maintain a control effect during the whole take-off process. In the case of SEA shown in Fig. \ref{fig:take-off-exp} (c), because the pitch is fully controlled by the swashplateless mechanism, which is not affected by the near-ground airflow, the pitch error during take-off is minimal (1.4$^{\circ}$).

\begin{figure}[t]
    \begin{center}
        {\includegraphics[width=1\columnwidth]{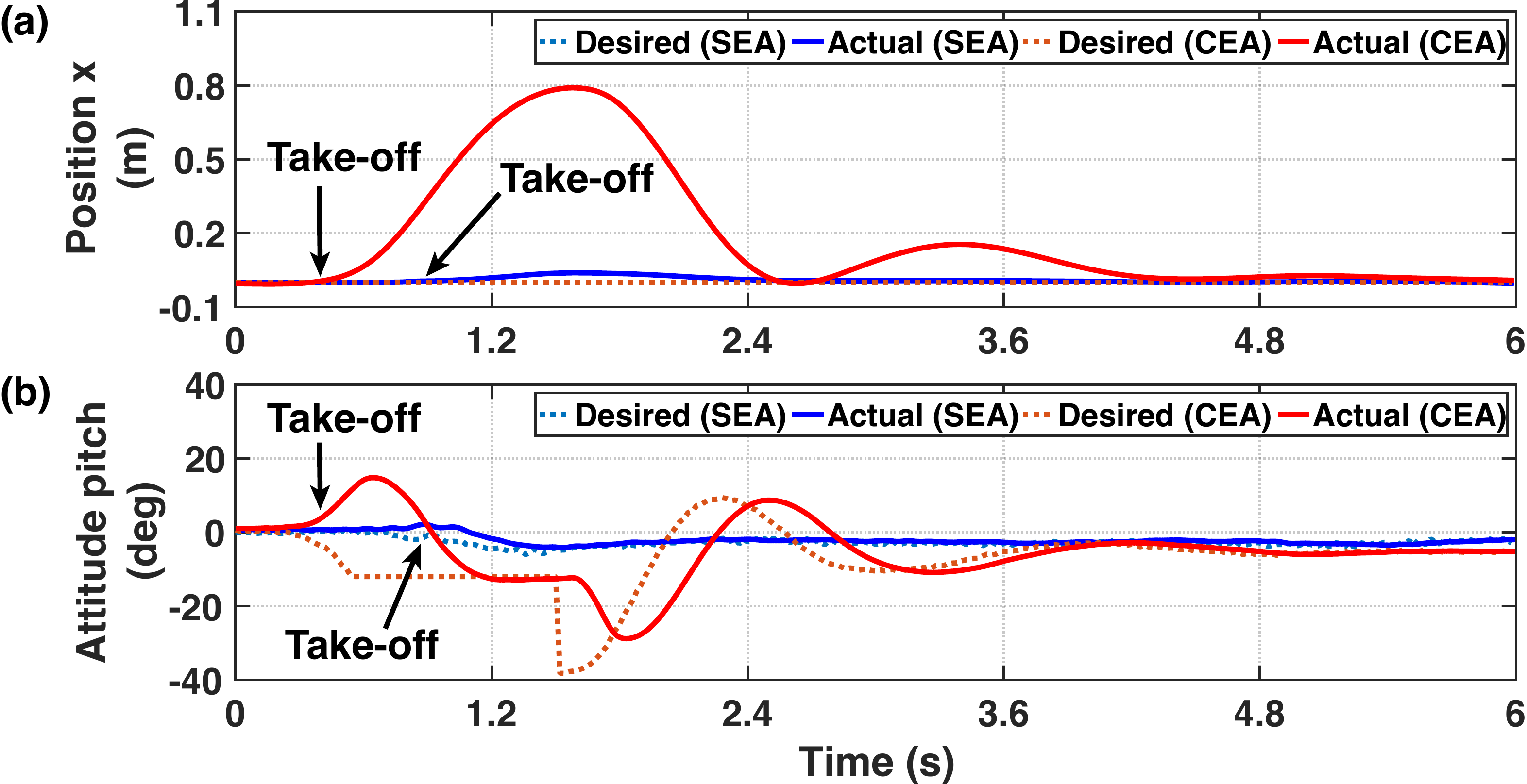}}
    \end{center}
    \vspace{-0.3cm}
    \caption{Comparison of $x$-position error and pitch angle error during take-off in position control mode with different actuation approach.}
    \label{fig:take-off-pos}
     % \vspace{-0.5cm}
\end{figure}

By turning the position controller off, one can manually control the throttle and allow the UAV to take-off quickly. In such a case, the pitch error can be less than 10$^{\circ}$ and does not cause significant disadvantages. However, when the UAV takes off with position control, the throttle is decided by the calculation of position controller. If a smooth take-off is expected, the take-off throttle may increase gradually, resulting in a significant pitch angle and position error in the CEA. Figure \ref{fig:take-off-pos} shows the UAV's position and attitude during take-off in position control mode. A significant position error (0.79 m) is caused by the large pitch error (14.7$^{\circ}$) when using the CEA. For the SEA, the position error is only 0.04 m, which is merely 5.1\% of the CEA. Therefore, the take-off performance of the SEA control is much better than that of the CEA, leading to a smooth take-off even when the UAV is close to the ground, no matter the position controller is turned on or off.

\subsection{Comparison of Trajectory Tracking Performance}

\begin{figure}[t]
    \begin{center}
        {\includegraphics[width=1\columnwidth]{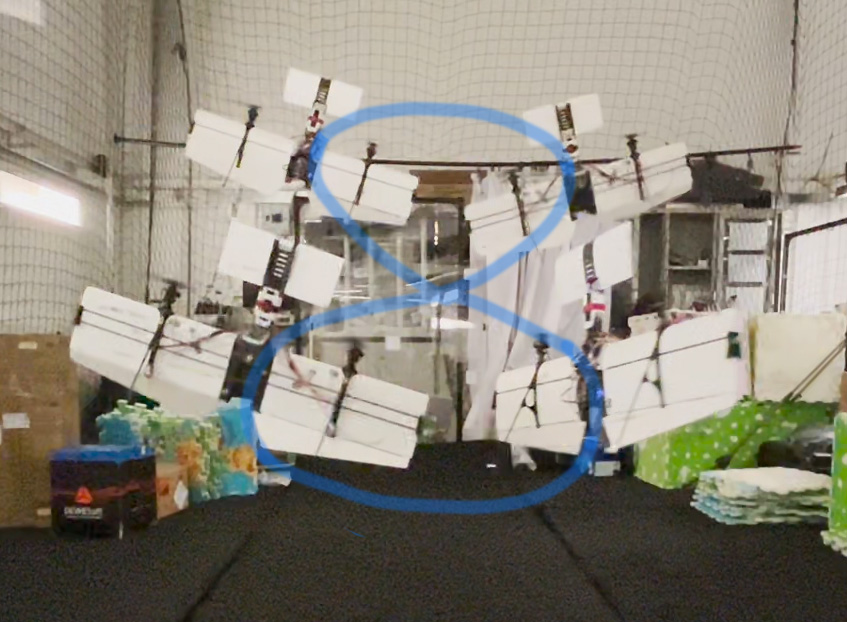}}
    \end{center}
    \vspace{-0.3cm}
    \caption{The overlaid snapshots of Hong He when it tracks a 3D figure-of-eight trajectory with continuous yaw rotation.}
    \label{fig:eight_tracking_3D_snap}
    % \vspace{-0.5cm}
\end{figure}

\begin{figure}[t]
    \begin{center}
        {\includegraphics[width=1\columnwidth]{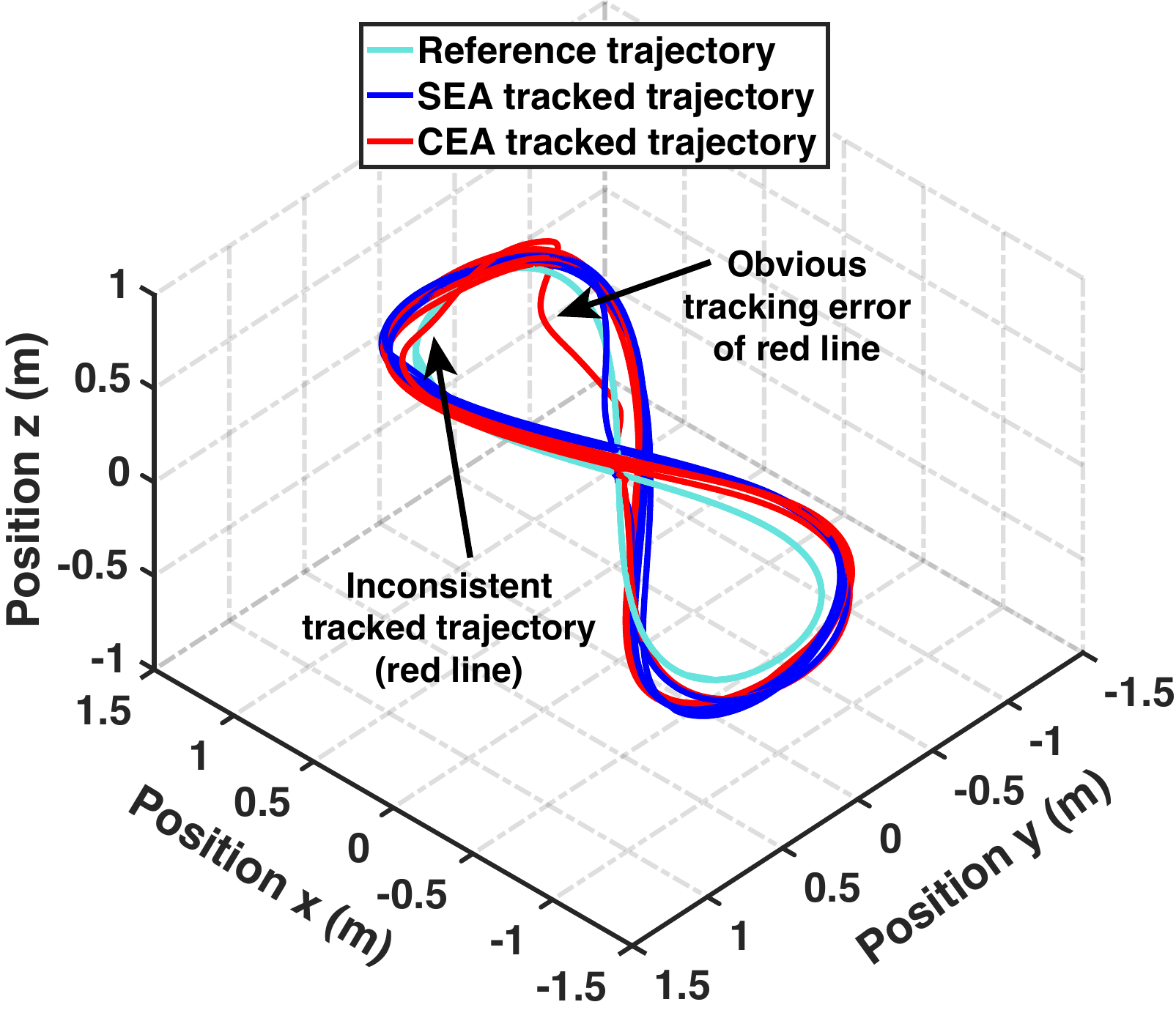}}
    \end{center}
    \vspace{-0.3cm}
    \caption{The reference trajectory and tracked trajectory in the trajectory tracking experiment.}
    \label{fig:eight_tracking_3D}
    % \vspace{-0.5cm}
\end{figure}

\begin{figure}[t]
    \begin{center}
        {\includegraphics[width=0.9\columnwidth]{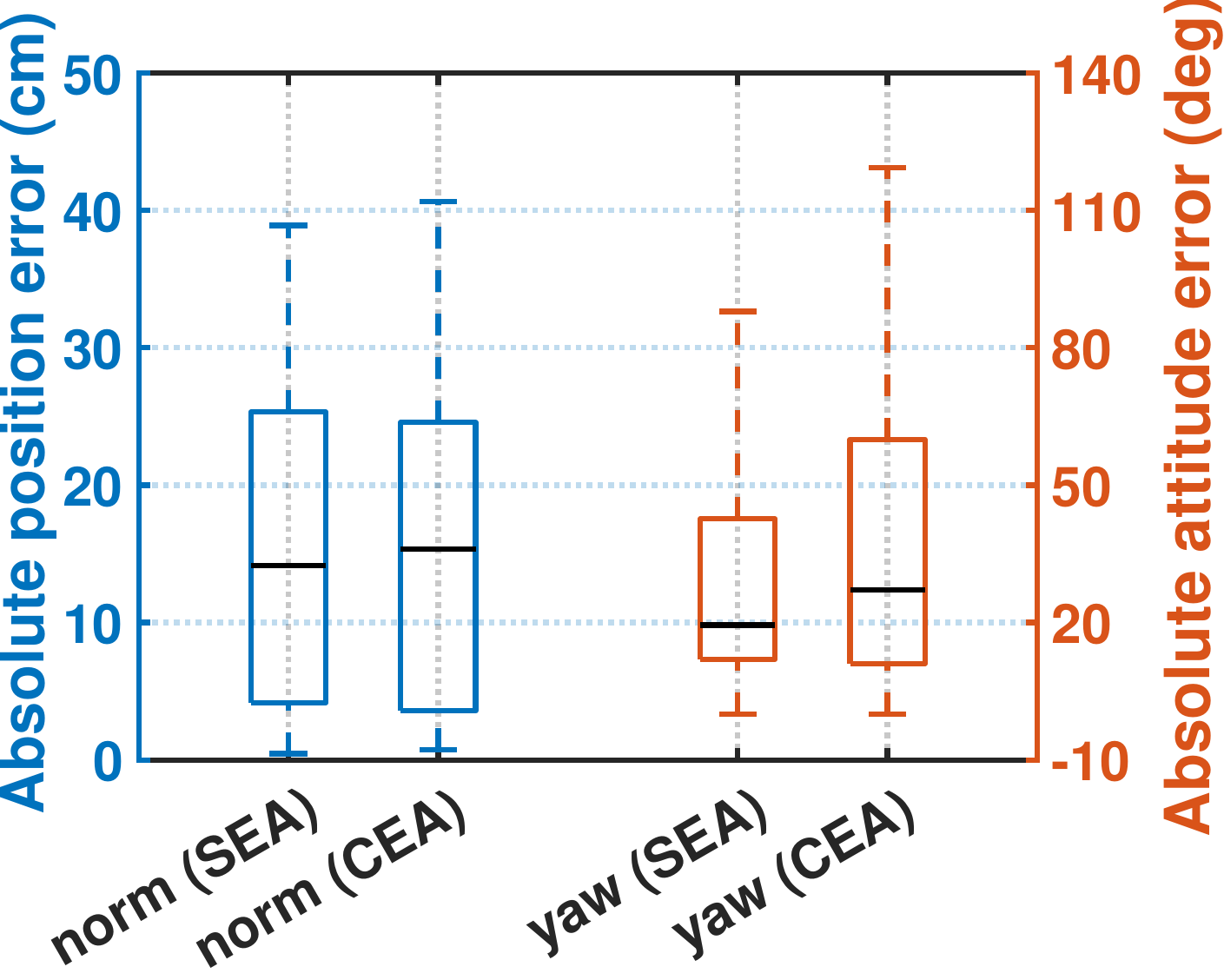}}
    \end{center}
    \vspace{-0.3cm}
    \caption{The errors of position norm and yaw angle during tracking the 3D figure-of-eight trajectory. The central mark indicates the median, and the bottom and top edges of the box represent the 25th and 75th percentiles, respectively. The whiskers extend to the maximum and minimum value.}
    \label{fig:eight_tracking_boxplot}
    % \vspace{-0.5cm}
\end{figure}

The tracking performance of both SEA and CEA are compared by conducting a figure-of-eight trajectory tracking experiment, as shown in Fig. \ref{fig:eight_tracking_3D_snap}. The length and width of the figure-of-eight trajectory are 2 m and 1 m, respectively, and the trajectory is executed for four cycles. The trajectory period is 5 s, but in the first and last cycles, the trajectory period is extended to 7.5 s to achieve smooth acceleration and deceleration. During tracking, the yaw angle is commanded to maintain the body $x$-axis same as the horizontal component of the velocity direction, which can generate more attitude control effort to evaluate the differences between SEA and CEA. The results in Fig. \ref{fig:eight_tracking_3D} show that the tracked trajectory of SEA is more consistent than that of CEA. The absolute errors of position norm and attitude yaw during tracking are shown in Fig. \ref{fig:eight_tracking_boxplot}.
For the absolute error of position norm, the median value of SEA and CEA are 14.17 cm and 15.35 cm, respectively, and the maximum values for both are 38.91 cm and 40.64 cm, respectively, indicating that the SEA has a slightly better performance than the CEA in the position control. For the absolute error of attitude yaw, the median value of SEA and CEA were 19.5$^{\circ}$ and 27.1$^{\circ}$, respectively, and the maximum value for both were 87.9$^{\circ}$ and 119.3$^{\circ}$, respectively, indicating that the SEA has relatively obvious advantage in yaw control than the CEA during the fast trajectory tracking.

\subsection{Comparison of Disturbance Rejection}

Three experiments are performed to evaluate the disturbance rejection performance of the SEA and the CEA: (i) hovering under balanced wind disturbance, (ii) forward position steps under unbalanced wind disturbance, and (iii) lateral position steps under unbalanced wind disturbance. The position information is provided by a motion capture system (Vicon).

\subsubsection{Hover under balanced wind disturbance}

\begin{figure}[t]
    \begin{center}
        {\includegraphics[width=1\columnwidth]{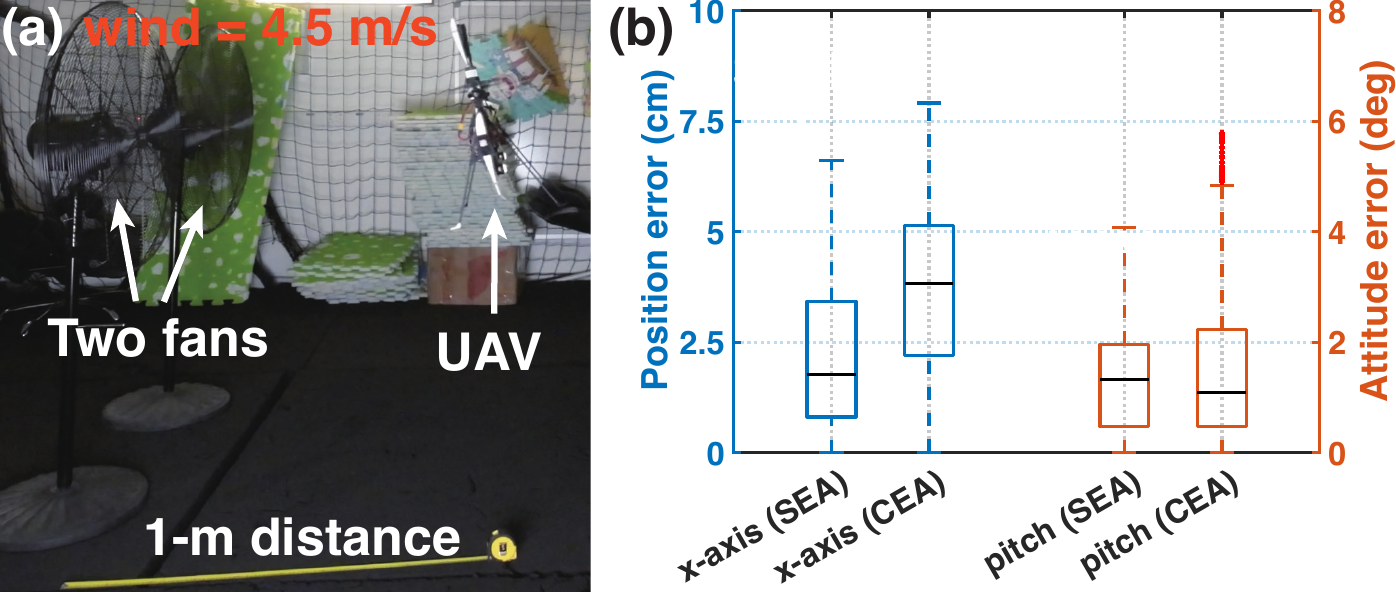}}
    \end{center}
    \vspace{-0.3cm}
    \caption{Hover under balanced wind disturbance. (a) Experimental setup for the disturbance rejection test of fan wind. (b) errors of $x$-axis position and pitch angle of SEA and CEA during a balanced wind disturbance.}
    \label{fig:dis_exp_setup}
    % \vspace{-0.4cm}
\end{figure}

As shown in Fig. \ref{fig:dis_exp_setup} (a), two fans are placed in parallel and located 1 m from the UAV's hover position. The fans are off initially and are then turned on to produce a wind gust of about 4.5 m/s in this distance. After the UAV reaches a stable pose, the fans are turned off to cancel the wind disturbance. The results are shown in Fig. \ref{fig:dis_exp_setup} (b). Since the wind disturbance is applied on the body $x$-axis, we can only focus on the errors on $x$-axis position and pitch angle. For the $x$-axis position, both the median value and maximum value of SEA are smaller than that of CEA, which are 1.77 cm versus 3.83 cm and 6.60 cm versus 7.92 cm, respectively. The results show the SEA can maintain the UAV's position better. For the pitch angle errors, although the CEA has slightly smaller median values than the SEA (i.e., 1.09$^{\circ}$ versus 1.32$^{\circ}$), the maximum value of CEA are relatively larger (i.e., 5.78$^{\circ}$ versus 4.08$^{\circ}$). This shows that both actuation approaches have good pitch angle control performance under wind disturbance but the SEA can realize a smaller error band.

\subsubsection{Forward position steps under unbalanced wind disturbance}

Maintaining good response in the presence of external disturbances is more difficult than without disturbance, as the actuators need to actively suppress the disturbance, leading to reduced actuator margins for tracking controller commands. To verify actual performance under disturbances, a position step response in the $x$ direction is performed with a fan wind disturbance being applied to a single side of the main wing.

\begin{figure}[t]
    \begin{center}
        {\includegraphics[width=1\columnwidth]{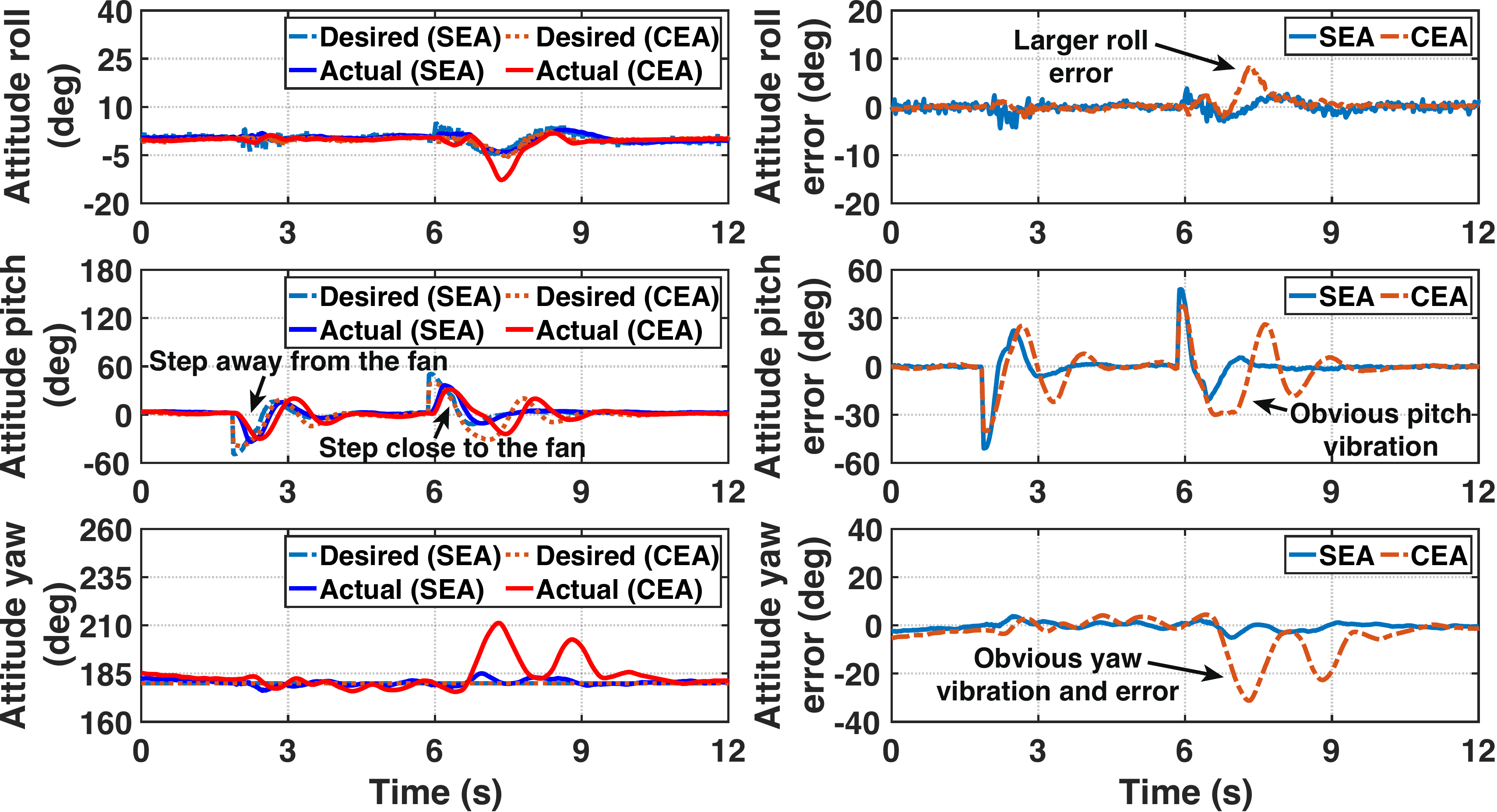}}
    \end{center}
    \vspace{-0.3cm}
    \caption{Attitude responses (left panel) and their corresponding errors (right panel) when applied a step command in body $x$ direction under an unbalanced wind disturbance on the main wing.}
    \label{fig:yaw_wind_step_att}
    % \vspace{-0.3cm}
\end{figure}

In the experiment, the UAV first hovers 1 m from a fan that has been turned on. Then, the UAV steps 1 m in the same direction as the wind airflow and finally steps back to the original hover position. Since the wind area of the UAV is large, we pay more attention to the attitude response during the wing disturbance. The results of attitude response is shown in Fig. \ref{fig:yaw_wind_step_att}. When the UAV steps away from the fan, the wind disturbance decreases. In this condition, and attitude response of the two actuation approaches are similar. However, when the UAV steps back to the original hover position, the wind disturbance increases. Because the elevons of CEA need to deflect an angle to generate control moment both in pitch and yaw directions, they cannot provide sufficient moment to resist the wind disturbance on these two directions. Consequently, when using CEA, obvious oscillation occurred in the converging process of pitch angle and significant shaking in yaw angle (maximum error of 31.2$^{\circ}$). Since large attitude error occurred in pitch and yaw, the wind disturbance affects roll more easily due to the changing of the windward side, leading to a maximum error of 8.2$^{\circ}$. In contrast, no oscillation or obvious error occurs in the attitude response of the SEA during the entire process, indicating that it can achieve more actuator margin and maintain good attitude control performance even with external disturbance.

\begin{figure}[t]
    \begin{center}
        {\includegraphics[width=1\columnwidth]{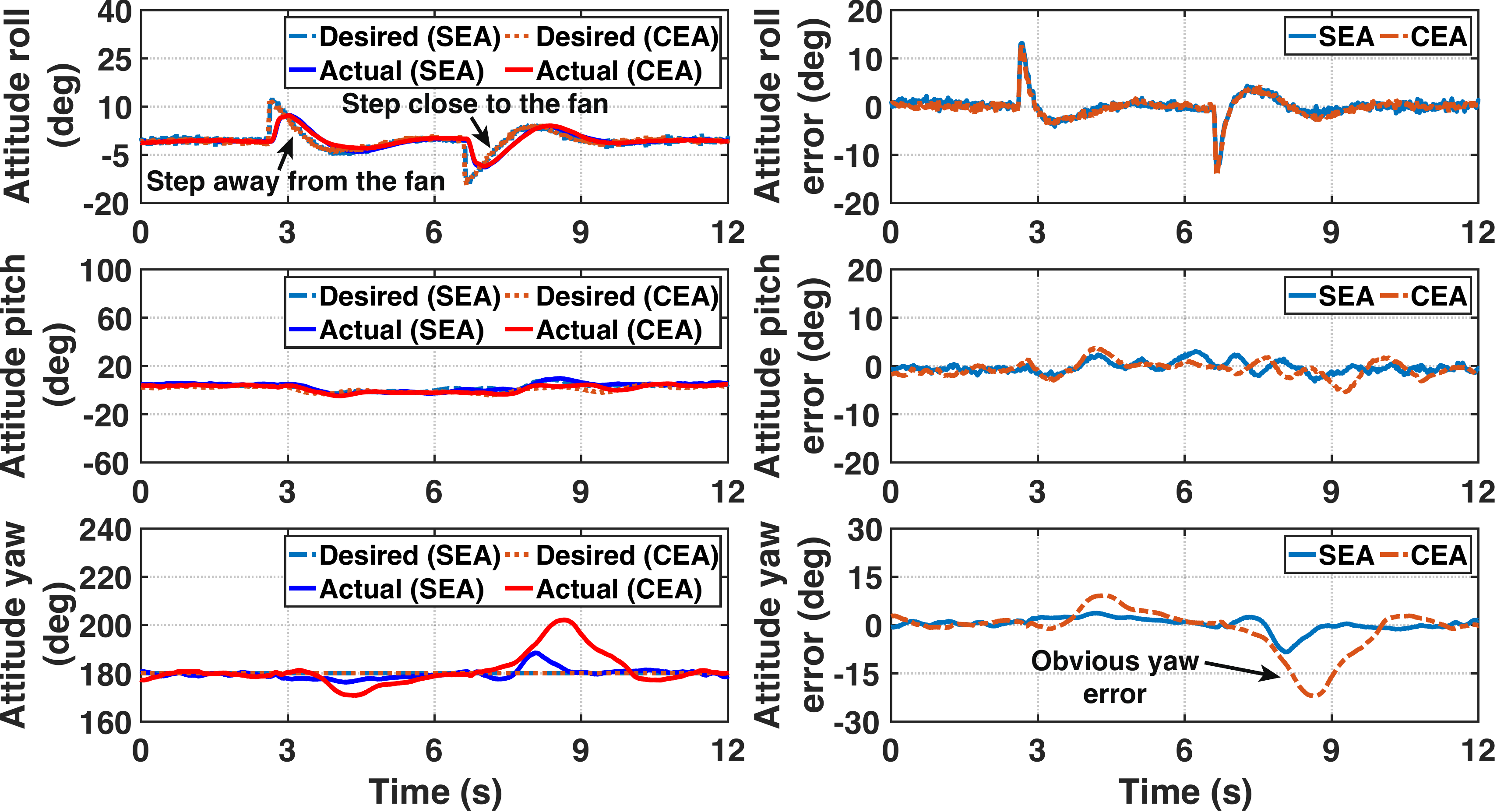}}
    \end{center}
    \vspace{-0.3cm}
    \caption{Attitude responses (left panel) and their corresponding errors (right panel) when applied a step command in body $y$ direction under an unbalanced wind disturbance on the main wing.}
    \label{fig:yaw_wind_roll_step_att}
    % \vspace{-0.2cm}
\end{figure}

\subsubsection{Lateral position steps under unbalanced wind disturbance}
In addition to the forward and backward position step experiment in the body $x$ direction, we also conducted a lateral position step experiment in the body $y$ direction. The experiment setup is same as before. The UAV initially hovered 1 m away from a fan that had been turned on, then stepped 1 m away from the fan in the direction which is orthogonal to the direction of the wind airflow, and finally stepped back to the original hover position. The attitude response results of this experiment are shown in Fig. \ref{fig:yaw_wind_roll_step_att}.

When the UAV stepped away from the fan, the wind disturbance decreased. In this condition, the attitude control of the two actuation approaches are still similar. However, when the UAV stepped back to the original hover position, the wind disturbance increased. Similar to the forward step experiment, the CEA showed obviously larger yaw angle shaking (maximum error of 22.1$^{\circ}$) than the SEA (maximum error of 8.4$^{\circ}$), further verifying that the SEA has better disturbance rejection performance than the CEA.

\subsection{Transition and Fixed-wing Mode Flight}

\begin{figure}[t]
    \begin{center}
        {\includegraphics[width=1\columnwidth]{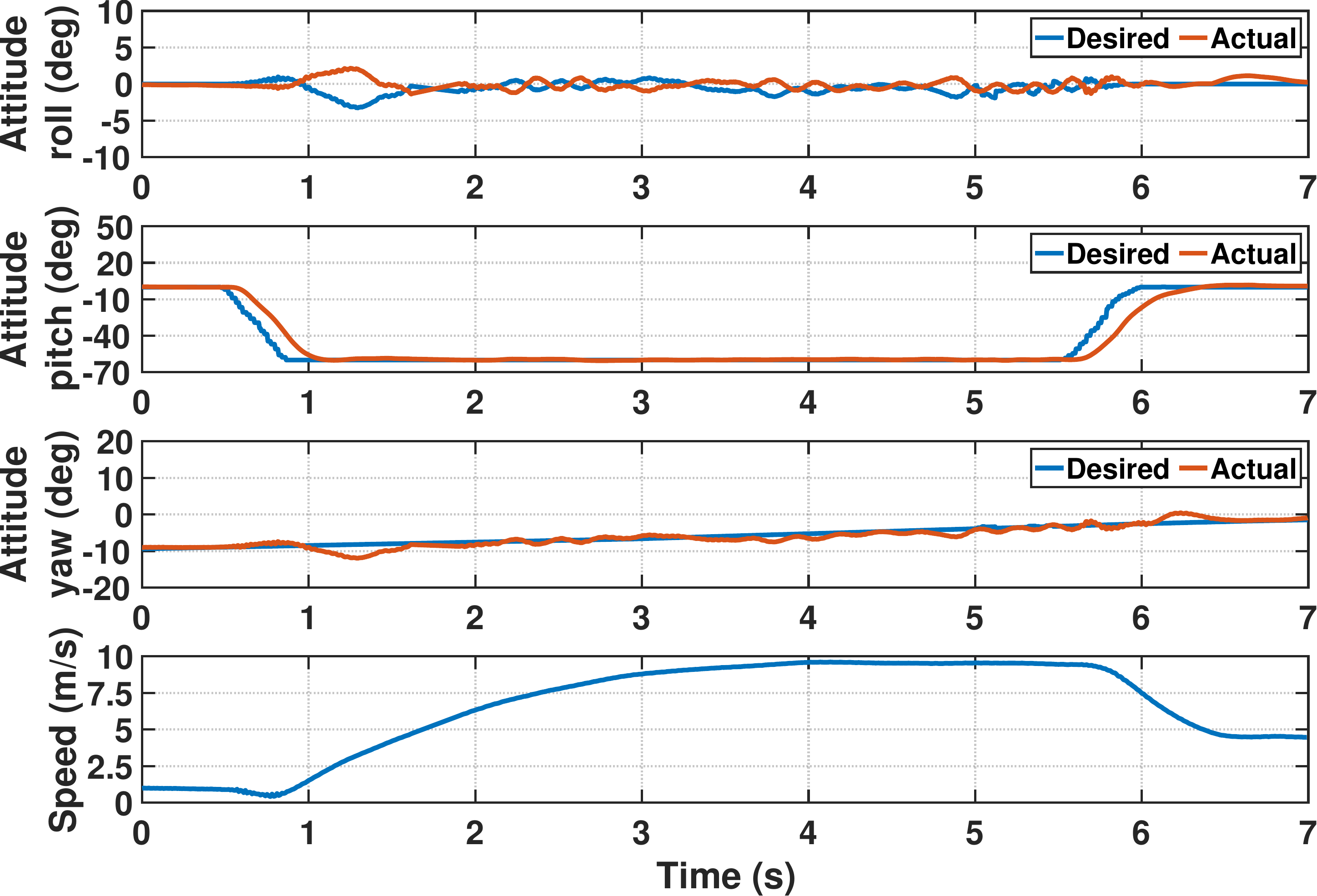}}
    \end{center}
    \vspace{-0.3cm}
    \caption{Desired attitude, actual attitude, and flight speed during transition and fixed-wing mode flight using SEA.}
    \label{fig:tran_att_vel}
    % \vspace{-0.5cm}
\end{figure}

We also conducted experiments to verify the performance of the proposed SEA in the fixed-wing mode flight as shown in Fig. \ref{fig:cover} (b). The results are shown in Fig. \ref{fig:tran_att_vel}. The definition of UAV coordinate frame in the fixed-wing mode is the same as that of the multi-rotor mode. The pitch angle of fixed-wing mode is set at -65$^{\circ}$ (i.e., the angle between main wing and horizontal plane is 35$^{\circ}$). During the transition, the attitude controller tracked the desired attitude accurately without any overshoot, showing that the swashplateless mechanism has a fast response and can generate sufficient moment. In the fixed-wing mode flight, the UAV accelerated continuously and finally reached a speed of 9.6 m/s. The pitch angle had no vibration and always remained at 65$^{\circ}$, indicating that the swashplateless mechanism can work well in the fixed-wing mode. However, the roll and yaw, which are controlled by the motors and elevons, respectively, had slight vibrations around the desired values. The maximum errors in roll and yaw appeared when the UAV just completed the transition process, which may have been caused by the airflow disturbance. Nonetheless, these errors were not obvious, only 5.3$^{\circ}$ in roll and 3.8$^{\circ}$ in yaw. It is should be noted that the CEA-based fixed-wing mode flight of a VTOL UAV has been verified previously. Hence, we did not present a quantitative comparison experiment here and only verify the feasibility of the fixed-wing mode flight based on SEA.

\section{Conclusion}

In this paper, an actuation approach called SEA is proposed for dual-rotor VTOL UAV to decouple pitch and yaw control and improve take-off and disturbance rejection performance. The proposed SEA-based UAV showed reduced pitch and position errors during take-off compared to the CEA-based UAV, which had noticeable errors due to ground-distorted airflow. The control performance of both SEA and CEA are evaluated by tracking a 3D figure-of-eight trajectory with continuous yaw angle rotation, showing that the SEA has less error both in position and yaw angle. Disturbance rejection performance was evaluated by the experiment of hovering in the balanced wind gust produced by two fans. The SEA exhibited better performance in position and pitch angle, indicating that SEA is more robust in an environment with wind gust. Step response experiments under unbalanced wind disturbance showed that the SEA outperformed the CEA with obviously lower attitude errors. These experiments validate that the SEA can mitigate actuator saturation by decoupling the actuation of pitch and yaw, and improve the performances of both tracking control and disturbance rejection. Finally, we validate the SEA in the transition process and fixed-wing mode flight in an outdoor environment, demonstrating its capability to maintain a stable attitude for a VTOL UAV in the presence of high-speed incoming airflow.

\addtolength{\textheight}{-12cm}



\section*{ACKNOWLEDGMENT}
This work is supported by the Grants Committee Early Career Scheme of The University of Hong Kong under Project 27202219, General Research Fund of Hong Kong under project 17206920, and a DJI research donation.

\bibliography{citation}
\bibliographystyle{IEEEtran}

\end{document}